\documentclass{article}

     \usepackage[preprint]{tackling_climate_workshop_style}

\usepackage[utf8]{inputenc} 
\usepackage[T1]{fontenc}    
\usepackage{hyperref}       
\usepackage{url}            
\usepackage{booktabs}       
\usepackage{amsfonts}       
\usepackage{nicefrac}       
\usepackage{microtype}      
\usepackage{graphicx}

\title{AI for Agriculture: the Comparison of Semantic Segmentation Methods for Crop Mapping with Sentinel-2 Imagery}

\author{%
  Irina Korotkova\thanks{https://www.deepplanet.ai} \\
  DeepPlanet\\
  London, UK, EC2A 2BB\\
  \texttt{irina@deepplanet.ai} \\
  \And 
  Natalia Efremova\\
  Queen Mary University\\
  London, UK \\
  \texttt{n.efremova@qmul.ac.uk} \\
}

\begin{document}

\maketitle

\begin{abstract}
Crop mapping is one of the most common tasks in  artificial intelligence for agriculture due to higher food demands from a growing population and increased awareness of climate change.  In case of vineyards, the texture is very important for crop segmentation: with higher resolution satellite imagery the texture is easily detected by majority of state-of-the-art algorithms. However, this task becomes increasingly more difficult as the resolution of satellite imagery decreases and the information about the texture becomes unavailable. In this paper we aim to explore the main machine learning methods that can be used with freely available satellite imagery and discuss how and when they can be applied for vineyard segmentation problem. We assess the effectiveness of various widely-used machine learning techniques and offer guidance on selecting the most suitable model for specific scenarios.
\end{abstract}

\section{Introduction}

Agriculture is both one of the sectors most susceptible to climate change and a significant contributor to it [1]. Therefore, it is essential to consider both mitigation and adaptation strategies, as well as transforming agricultural practices to promote sustainability and resilience [2]. A key objective of application of artificial intelligence (AI) and satellite imagery in agricultural setting is to develop more reliable and scalable methods for monitoring global crop conditions promptly and transparently, while also exploring how we can adapt agriculture to mitigate the effects of climate change. Agricultural monitoring with earth observation data provides a timely and reliable way to access the state of the field or farm and the surrounding territories, used for gathering data and producing forecasts. Monitoring with satellite imagery and other remote sensing tools, such as drones, is becoming mainstream, since it provides rapid precision data across the entire globe. Computer vision and signal processing techniques play crucial role in extracting meaningful information from raw satellite data. Growing adoption of AI and machine learning (ML) tools has significantly influenced the expansion Earth Observation (EO) and remote sensing to agricultural management. These advanced techniques are employed throughout the entire data processing cycle, encompassing tasks such as data compression, transmission, image recognition, and forecasting environmental factors like land cover, land use, biomass, and more [3].

The key areas of research in AI for agriculture include crop mapping [4], crop type mapping [4], field boundary delineation [5], yield estimation [7], and pests and disease detection [8]. These problems align with various machine learning tasks and problem types. For instance, crop mapping involves binary classification of pixels as either crop or non-crop. Crop type mapping requires multi-class classification or one-versus-all binary classification. Field boundary delineation entails image segmentation and providing the boundaries of agricultural fields as an output. 

\begin{figure}
  \centering
  \includegraphics[width = 8cm]{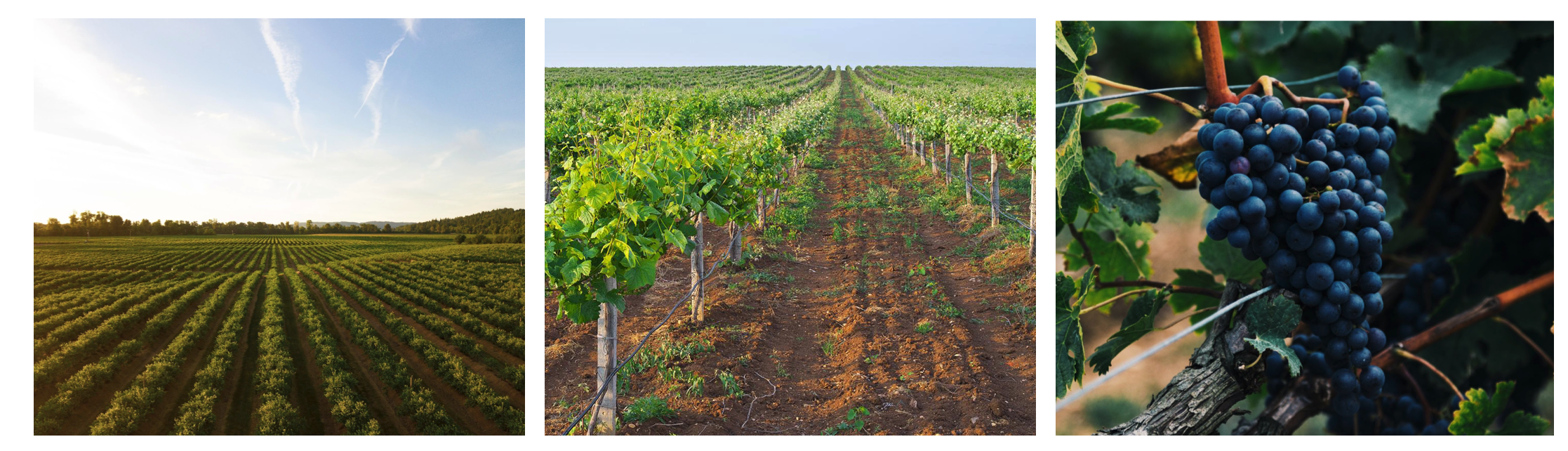}
  \caption{Vineyards have a distinct texture due to their rows, which is visible in very high-resolution satellite images. However, these images are often not publicly available, infrequent, and costly. For vineyard segmentation task, we use the spectral features of Sentinel-2 images, which are captured every 5-6 days.}
  \label{fig:photos}
\end{figure}

Although crop mapping in general [4-6] and binary vineyard segmentation in particular [9] are quite popular research topics, there are few works that look at vineyard segmentation with freely available satellite imagery. The importance using publicly available data can be explained by looking at the main stakeholders in this problem in agriculture: the recipients of agricultural and food security information include farmers and policy makers. Farmers require data on optimal planting times, crop performance, threats to production, soil moisture, potential productivity, and crop suitability. Policy makers seek insights into these same areas, along with guidance on intervention strategies, measuring policy impacts, and adapting to climate change projections. To be practically useful, crop mapping needs to be accurate, efficient, fast and cheap, both financially and computationally since it needs to be executed every growing season for policy makers' goals and weekly for farmers' goals.

In case of vineyards, the texture is very important for crop segmentation (Fig.\ref{fig:photos}): on higher resolution satellite images (between 3 m per pixel and 0.5 m per pixel) the texture will be easily detected by majority of state-of the-art segmentation algorithms, even with a few spectral bands [9]. However, this task becomes increasingly more difficult as the resolution of satellite imagery decreases and the information about the texture becomes unavailable. In this paper we aim to explore the main ML methods that can be used with freely available satellite imagery and discuss how and when they should be applied for vineyard detection.  
 
\section{Methods}
\label{sec:methods}
\subsection{Data}
\label{subsec:data}
Twelve  spectral bands from the Sentinel-2A sensor were obtained from multiple regions within Australia. The images were selected for summer 2021-2022. These bands were combined with the Normalized Difference Vegetation Index (NDVI). NDVI is the most commonly used vegetation index that takes the difference between the near infrared and red reflectance divided by their sum.  Low values of NDVI generally correspond to barren areas of rock, sand exposed soils, water, snow or clouds, increasing NDVI values indicate green or vegetation, including forests, crop lands, and wetlands. NDVI is efficient and simple index to identify vegetated areas and their conditions, it also reduces sun angle shadow and topographic variation effects. The acquired images were pre-processed with min-max normalization, so that their values fell within the [0, 1] range. We manually created corresponding binary masks over the three main wine regions in Australia.

Our approach involved the random selection of 20,000 image patches, each sized at 96 pixels. The choice of patch size was dictated by the average size of the block on the region of interest. It might differ for different crops and block sizes.These patches were partitioned into three datasets: 80\% for training, 10\% for validation, and the remaining 10\% for testing purposes.

\subsection{Model architecture}
\label{subsec:arch}
We experimented with different classic machine learning architectures, including decision trees, logistic regression, random forest (Table \ref{table:ml}) and deep learning architectures:  U-Net [10], UNet++ [11] ,Res-UNet [12], and ModSegNet [9] (Table \ref{table:dl}). We explored the range of parameters, including learning rates, dropout rates and the number of downsamplind-upsampling stages. U-Net is a convolutional neural network architecture designed for image segmentation tasks, it has an encoder-decoder architecture with pathways between stages, transferring feature masks from the beginning to the end of network. UNet is well-known for its effectiveness in segmentation tasks. Our experiments included various depths of U-Net model: 4-6 downsampling-upsampling stages, 16 or 32 filters, and various dropout rates. Res-UNet uses key features of both UNet ans ResNet [13] models, where deep U-shape architecture combines with residual connections, preventing vanishing and exploding of gradient. We tested a number of models with a depth of 4 or 5, 16 or 32 filter number, 0.0001-0.0005 learning rates, and a range of dropouts. UNet++ is a nested version of U-Net model with dense skip connections. These blocks allow to reduce semantic difference between contracting and expanding paths of the model. We compared a range of filters with other parameters being fixed. ModSegNet uses the same pooling indices in the same stages of the model together with feature masks. We tested different number of filters for 4 stages architecture with dropout=0.5 at the deepest layers.

\begin{figure}
  \centering
  \includegraphics[width = 12cm]{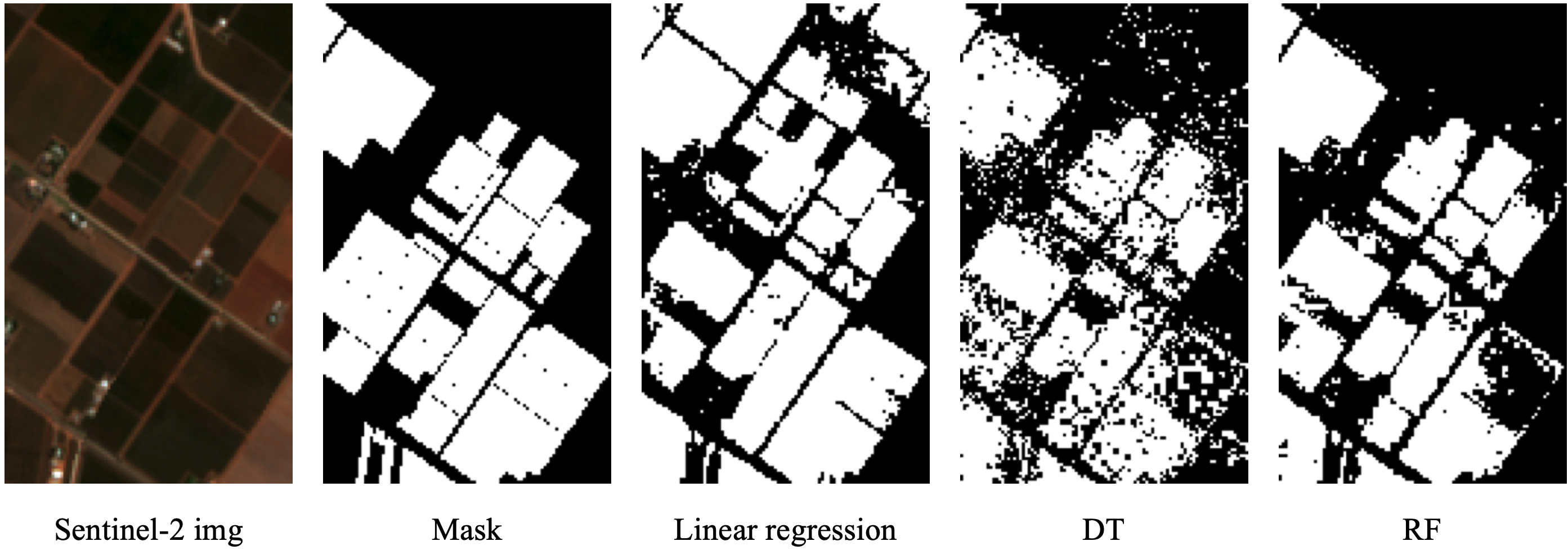}
  \caption{Left to right: original Sentinel-2 image,  binary mask, linear regression model output, decision trees model output, random forest output.}
  \label{fig:ml}
\end{figure}

\begin{figure}
  \centering
  \includegraphics[width = 12cm]{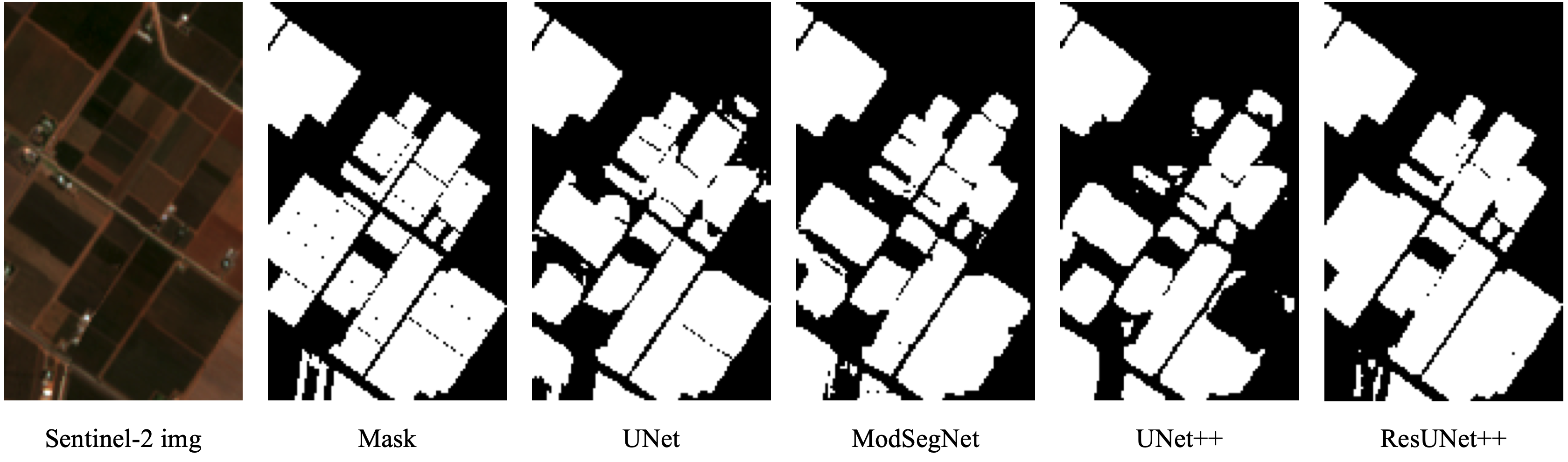}
  \caption{Left to right: original Sentinel-2 image,  binary mask, UNet model output, Modsegnet model output, UNet++ model output, ResUnet model output.}
  \label{fig:dl}
\end{figure}

\subsection{Experiments}
\label{subsec:experiments}
We conducted a series of experiments, while keeping the following parameters the same for all deep learning models: batch size 128, loss: dice loss.
While training UNet models, we tested different number of filters (16/32), depths(4-6), and dropout rates(0-0.3). Larger dropout or deeper architecture did not always provide higher dice scores, but the performance of the model with 32 filters typically surpassed the one with 16 filters. In ResUNet architectures, we did no find  correlation between dice and depth(4-5), number of filters(16/32), or dropout(0-0.3). However, the increase of learning rate from 0.0001 to 0.0005 gave us increase of dice by 0.01-0.02. In UNet++ the larger number of filters (depth=4, no dropout) did not provide any improvement. In ModSegNet, 64 filters slightly improved the result, compared to 32 filters. 

\begin{table}
  \caption{ML methods}
  \label{table:ml}
  \centering
  \begin{tabular}{ll}
    \toprule   
    Model     & Dice coef      \\
    \midrule
    Decision Tree & 	0.8399     \\
    Logistic Regression & 0.8155    \\
    \textbf{Random Forest }  & \textbf{0.8985} \\
    \bottomrule
  \end{tabular}
\end{table}		
			
\begin{table}
  \caption{DL methods}
  \label{table:dl}
  \centering
  \begin{tabular}{llll}
    \toprule
    Model     & Learning rate	&Dropout rate	&Dice coefficient \\
    \midrule
    UNet 	&0.001	&0	&0.92\\
    \textbf{ResUNet} 	&\textbf{0.0005}	 &\textbf{0}  &0.\textbf{93}\\
    UNet++ 	&0.001	&0	&0.92\\
    ModSegNet 	&0.001	&0.5  &0.92\\	
    \bottomrule
  \end{tabular}
\end{table}

\begin{figure}
  \centering
  \includegraphics[width = 8cm]{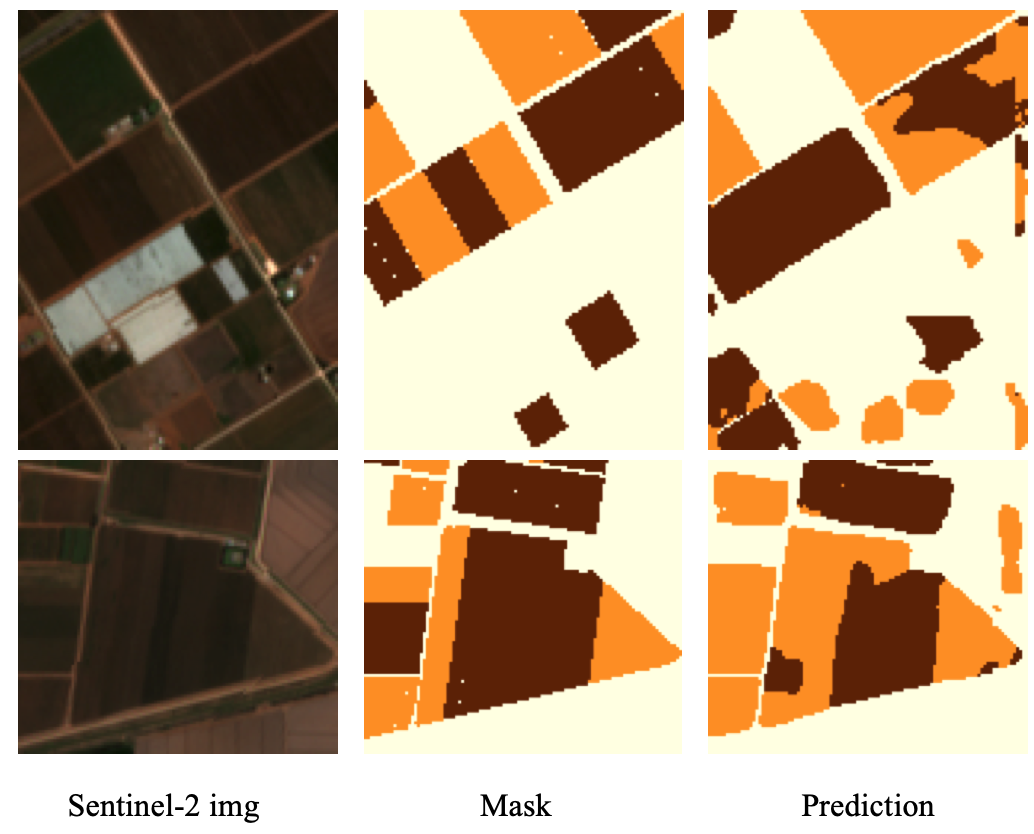}
  \caption{Left to right: original Sentinel-2 images,  multi-class mask, UNet model output.}
  \label{fig:red_white}
\end{figure}

\section{Results and discussion}
\label{sec:results}
The results demonstrate that the differences between the applied models' performance is relatively small. Visual assessment of models' performance showed that many of the tested models can be used for agricultural purposes (Fig.\ref{fig:ml} and \ref{fig:dl}). The best performing model with Dice coefficient 0.93 was proven to be Res-Unet on Sentinel-2A product. The performance of the conventional ML methods was worse, but not significantly: the difference between the random forest model and ResUnet model was around 4\%. In more complex tasks, however, such as grape variety detection and distinction between the type of grapes, random forest model achieves better performance as it better suited to incorporate in-season variations between different types of crops. Our experiments suggest that semantic segmentation models perform well on 'easier' crop-type mapping tasks, such as distinction between the red and white grapes (Fig.\ref{fig:red_white}), on which U-net model demonstrated 80\% dice score, but decreases as we try to distinguish e.g. between the the red varieties or table and wine grapes. The reason for that is similarity in spectral characteristics in similar crops. However, models like random forest [14] or XGboost [15] would be more suitable to distinguish temporal differences in wine grape development, such as timing of the budburst, harvest dates etc.

\section*{References}

\small

[1] Mahowald, N.M.,\  Ward, D.S. \ Doney, S. C.  \ Hess, P.G. \ Randerson, J. T. \  (2017) { \it Are the impacts of land use on warming underestimated in climate policy?}, Environmental Research Letters 12 (9).

[2] Lee, H.\  Calvin, K \ Dasgupta, D \ et al. \ (2023) {\it Ar6 synthesis report: Climate change 2023}. Summary for Policymakers.

[3] Tuia, D. \  Schindler, K. \  Demir, B. et al. \ (2023) {\it Artificial intelligence to advance earth observation: a perspective}, arXiv preprint: 2305.08413 (2023).

[4] H. Kerner, \  G. Tseng, \  I. Becker-Reshef, \  C. Nakalembe, \  B. Barker, \  B. Munshell,\  M. Paliyam, \   M. Hosseini (2022) { \it Rapid response crop maps in data sparse regions}.

[5] S. Wang, \  S. Di Tommaso, \  J. M. Deines, \ D. B. Lobell \ (2020) {\it  Mapping twenty years of corn and soybean across the us midwest using the landsat archive}. Scientific Data 7 (1), 307.

[6] S. Wang, \ F. Waldner,\   D. B. Lobell \ (2022) {\it Unlocking large-scale crop field delineation in smallholder farming systems with transfer learning and weak supervision}. Remote Sensing 14 (22) 5738.

[7] Dado, W. T.,\ Deines, J. M.,\ Patel, R.,\ Liang, S. Z., \ \& Lobell, D. B. \ (2020). { \it High-resolution soybean yield mapping across the US Midwest using subfield harvester data}. Remote Sensing, 12(21), 3471.

[8] G. Silva,\   J. Tomlinson, \   N. Onkokesung, \   S. Sommer, \  L. Mrisho, \  J. Legg, \ I. P. Adams, \  Y. Gutierrez-Vazquez et al.\ (2021) {\it  Plant pest surveillance: From satellites to molecules}. Emerging topics in life sciences, 5 (2) 275–287.

[9] Barros, T.,\  Conde, P.,\  Gonçalves, G.,\  Premebida, C.,\  Monteiro, M.,\  Ferreira, C.S.S. \  Nunes, U.J. \ (2022). {\it Multispectral vineyard segmentation: A deep learning comparison study}. Computers and Electronics in Agriculture, 195, p.106782.

[10] Ronneberger, O.,\   Fischer, P., \   Brox, T. \ (2015). {\it U-net: Convolutional networks for biomedical image segmentation}. MICCAI 2015: 18th International Conference, Munich, Germany, October 5-9, 2015, Proceedings, Part III 18 (pp. 234-241). Springer International Publishing.

[11] Zhou, Z.,\   Rahman Siddiquee, M. M.,\   Tajbakhsh, N., \   Liang, J. \ (2018). { \it Unet++: A nested u-net architecture for medical image segmentation}. MICCAI 2018, Granada, Spain, September 20, 2018, Proceedings 4 (pp. 3-11). Springer International Publishing.

[12] Diakogiannis, F. I.,\  Waldner, F.,\ Caccetta, P.,\   Wu, C. \ (2020). { \it ResUNet-a: A deep learning framework for semantic segmentation of remotely sensed data}. ISPRS Journal of Photogrammetry and Remote Sensing, 162, 94-114.

[13] He, K., \ Zhang, X., \ Ren, S., \ Sun, J. \ (2016). {\it Deep residual learning for image recognition}. In Proceedings of the IEEE conference on computer vision and pattern recognition (pp. 770-778).

[14] Breiman, L. \ (2001). { \it Random forests}. Machine learning, 45, 5-32.

[15] Chen, T., \ \& Guestrin, C. \ (2016,). { \it Xgboost: A scalable tree boosting system}. In Proceedings of the 22nd acm sigkdd international conference on knowledge discovery and data mining (pp. 785-794).

\end{document}